\documentclass[letterpaper, 10 pt, conference]{ieeeconf}  

\IEEEoverridecommandlockouts                              
\usepackage{mathptmx} 
\usepackage{amsmath} 
\usepackage{amssymb}  
\usepackage[dvipdfmx]{graphicx}

\newcommand{\bm}[1]{\mbox{{\boldmath $#1$}}}

\title{\LARGE \bf
Tool Exchangeable Grasp/Assembly Planner
}

\author{Kensuke Harada$^{1,2}$, Kento Nakayama$^{1}$, Weiwei Wan$^{1,2}$, Kazuyuki Nagata$^{2}$\\
Natsuki Yamanobe$^{2}$, and Ixchel G. Ramirez-Alpizar$^{1}$
\thanks{$^{1}$Graduate School of Engineering Science, 
Osaka University, Toyonaka, Japan 
{\tt\small harada@sys.es.osaka-u.ac.jp}}
\thanks{$^{2}$Intelligent Systems Research Institute, 
National Inst. of Advanced Industrial Science and Technology, Tsukuba, Japan}
}

\begin{document}

\maketitle
\thispagestyle{empty}
\pagestyle{empty}


\begin{abstract}

This paper proposes a novel assembly planner for a manipulator which can simultaneously plan assembly sequence, robot motion, grasping configuration, and exchange of grippers. 
Our assembly planner assumes multiple grippers and can automatically selects a feasible one to assemble a part. 
For a given AND/OR graph of an assembly task, we consider generating the assembly graph from which assembly motion of a robot can be planned. The edges of the assembly graph are composed of three kinds of paths, i.e., transfer/assembly paths, transit paths and tool exchange paths. 
In this paper, we first explain the proposed method for planning assembly motion sequence including the function 
of gripper exchange. 
Finally, the effectiveness of the proposed method is confirmed through some numerical examples and a physical experiment. 
\end{abstract}

\section{INTRODUCTION}

\begin{figure}[t]
\centering
 	\includegraphics[width=78mm]{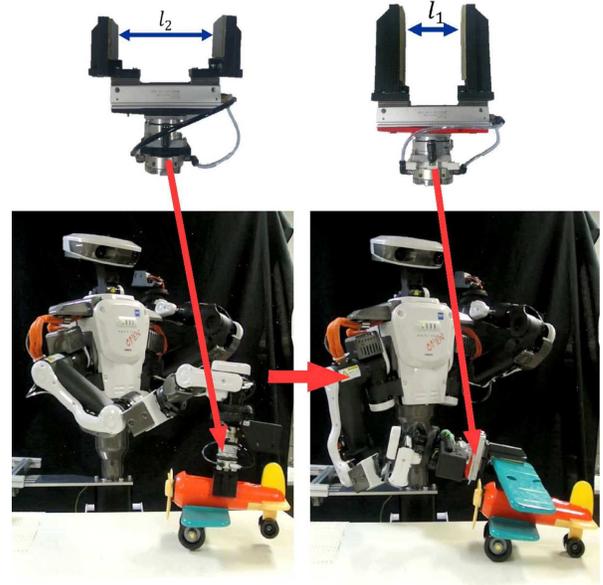}
	\caption{Industrial robots with a tool changer where a gripper was selected depending on each task }
	\label{fig.1}
\end{figure}

In factory environments, industrial robots are expected to assemble a product. 
During a robotic assembly process, robotic grippers have to firmly grasp a variety of parts with a variety of physical parameters such as shape, weight and friction coefficient. However, even if we design a gripper to firmly grasp a part, it is not always possible for the same gripper to firmly grasp the other parts with different physical parameters. 
To cope with this problem, a robotic manipulator used to assemble a product usually equips a tool exchanger at the wrist. 
By using a tool exchanger, we can selectively use a gripper from multiple candidates. 
As shown in Fig. \ref{fig.1}, we prepared two parallel grippers with different sized fingers. 
To assemble a toy airplane, a robot first grasps the body and places it on a table. 
Then, a robot grasps the wing and assembles it to the body. In this example, it is difficult for a robotic gripper to firmly grasp the wing by using the gripper used to grasp the body. 
Selection of a gripper is often more difficult and complex than this example since a robot has to select a suitable gripper from a set of two-fingered parallel jaw grippers, three-fingered grippers and suction grippers. 
So far, a gripper use to assemble a part has been selected based on the experience of human workers. On the other hand, this paper aims to construct a grasp/assembly planner which can automatically determine a gripper suitable for a given assembly task. 

The robotic assembly is a classical topic of robotics extensively researched by many researchers such as 
\cite{sanderson1,sanderson2,Kwak,Wilson,Heger,Thomas}. 
However, in most of the previous researches on assembly planners \cite{Kwak,Wilson,Heger}, 
grasping posture of a part was assumed to be known. 
While some researchers such as \cite{Simeon,Laumond,Vahrenkamp,Wan_RAL}
have proposed manipulation planners combined with grasp planners, 
it is relatively recently where assembly planners combined with grasp planners have proposed 
\cite{Thomas,Dogar,Wan1}. Hereafter, we call such assembly planner as the grasp/assembly planner. 
However, in spite of the fact that the tool exchanging capability is needed for actual assembly tasks, there has been no research on grasp/assembly planner taking the tool exchanging capability into consideration. 
We believe that this is the first trial on adding a function of automatically selecting a gripper to a grasp/assembly planner. 
In our previous research, we have proposed a dynamic regrasp graph \cite{Wan1,Wan_RAL} for solving 
a grasp/manipulation and grasp/assembly planning problems. 
On the other hand, this research newly assumes multiple grippers for the grasp/assembly planner proposed so far \cite{Wan1}. We show that, by using our proposed grasp/assembly planner, it becomes possible for automatically selecting a gripper from multiple candidates to assemble a part. 

The rest of the paper is organized as follows. After discussing previous works in Section 2, we show the definitions used in this research in Section 3. Section 4 formulates our proposed grasp/assembly planner. In Section 5, we confirm the effectiveness of our proposed method through a few numerical examples and a physical experiment where we assume two two-fingered grippers with different size. We show that, according to the shape and the size of a assembled part, our proposed planner can automatically select a feasible one and can complete an assembly task. 


\section{Definitions}

\begin{figure}[t]
\centering
 	\includegraphics[width=85mm]{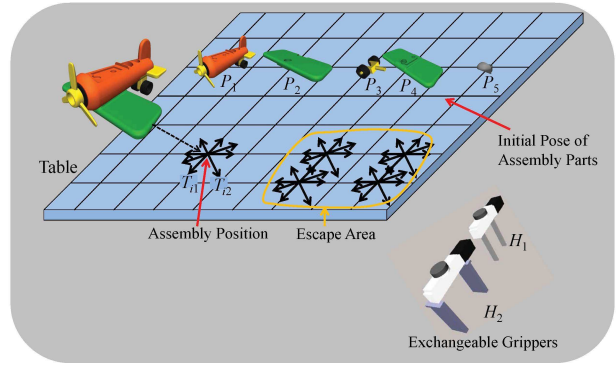}
	\caption{Definition of working area performing product assembly}
	\label{fig:graspPos-1}
\end{figure}

\begin{figure}[t]
\centering
 	\includegraphics[width=85mm]{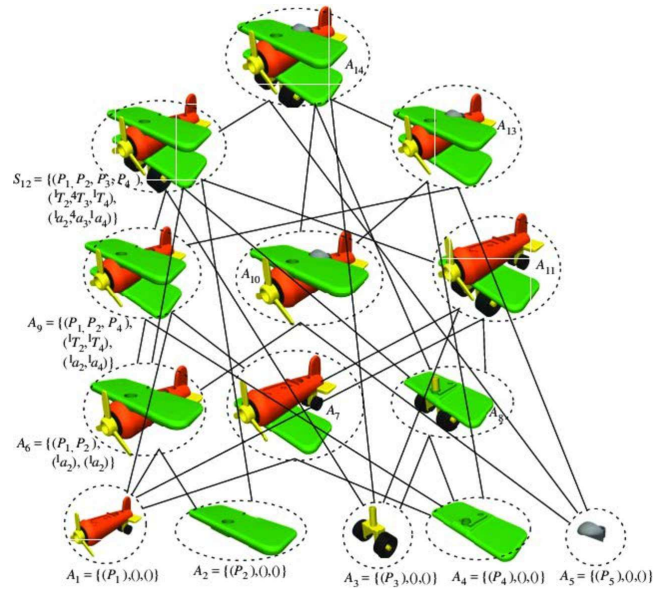}
	\caption{A part of AND/OR graph of a toy airplane}
	\label{fig:andor}
\end{figure}

Let us consider a product composed of $m$ parts $P=(P_1, \cdots, P_m)$ as shown in Fig. \ref{fig:graspPos-1}. 
Let $A=(A_1, \cdots, A_n)$ be the assembly of parts as shown in Fig. \ref{fig:andor} where $n \le \sum_{i=1}^m C_i^m$. 
For example, if the assembly $A_i$ is composed of the parts $P_u$, $P_v$ and $P_w$, it is defined as
\begin{equation}
A_i = \{ (P_u, P_v, P_w), (^u\bm{T}_v, ^u\bm{T}_w), (^u\bm{a}_v, ^u\bm{a}_w) \}
\end{equation}
where $^u\bm{T}_v$ denotes a 4$\times$4 homogenous matrix expressing the pose of the part $P_v$ relative to the part 
$P_u$, and $^u\bm{a}_v$ denotes a 3 dimensional unit vector expressing the approach direction of the part $P_v$ relative to the part $P_u$. On the other hand, if the assembly $A_i$ is composed only of the part $P_i$, it can be defined as follows:

\begin{equation}
A_i = \{ (P_i), (), () \},\ i=1,\cdots,m
\end{equation}

A product assembly is composed of a sequence of individual assembly tasks. 
Possible assembly sequences can be expressed by using the AND/OR graph $G(A,E)$ \cite{sanderson1} where it is composed of the assembly of parts $A$ as the vertices and the edges $E$ connecting them. 
An example of the AND/OR graph is shown in Fig. \ref{fig:andor}. 
Assembly sequence can be generated by searching for this graph. 

Let us consider a case where a robot performs a sequence of assembly tasks on a horizontally flat table. 
Let us consider discretizing the horizontal area of the table. 
We impose the following assumptions:
\begin{itemize}
\item[]{\bf A1}: A robot assembles a product by using a single arm. 
\item[]{\bf A2}: A robot performs an assembly task by once placing an assembly of parts at one of the grid points hereafter called the assembly point. 
\end{itemize}
Under these assumptions, a robotic gripper picks up an assembly of parts from the table and fit it to another assembly of parts placed at the assembly point. 

According to the assumption {\bf A1}, we further impose the following assumption:
\begin{itemize}
\item[]{\bf A3}: After finishing an individual assembly task, the assembly of parts is once moved to one of the grid points included in the escape area. 
\end{itemize}

\noindent
Let us consider preparing $h$ multi-fingered grippers as $H=(H_1, \cdots, H_h)$. 
In this research, we consider using a grasp planner such as \cite{Harada1,Harada2} to calculate a grasping posture of a part. 
We can use any multi-fingered grippers as far as a grasping posture can be calculated. 
For each pair of a grasped object and a gripper, 
we consider preparing a database of stable grasping postures. 
When a robot tries to actually grasp an object, we consider searching for the database to find a stable grasping posture. 
Let $G_{ij}=(G_{ij1}, \cdots, G_{ijk})$ be a database of grasping postures of the part $P_i$ grasped by the gripper $H_j$ where each element is composed of the wrist's pose with respect to coordinate system fixed to the part $P_i$ and joint angles of each finger. 

We additionally impose the following two assumptions:
\begin{itemize}
\item[]{\bf A4}: The AND/OR graph is given in advance of planning the assembly motion of a robot. 
\item[]{\bf A5}: Once a part is assembled, the assembly of parts will not be broken. 
\end{itemize}
As for the assumption {\bf A4}, since there have been a number of researches on automatically generating the AND/OR graph such as \cite{sanderson2,Kwak}, we can follow their research if we want to automatically generate the AND/OR graph. 

\section{Assembly/Grasp Planner}

This section details the grasp/assembly planner proposed in this research. 

\subsection{Placing Pose}

To obtain a set of stable placing postures of the assembly $A_i$, 
we first calculate its convex hull. 
We consider drawing a line including the assembly $A_i$'s CoG and perpendicular to one of the convex hull's facet. 
If this line passes through the facet, 

Let $T_{ik}$ be the homogenous matrix expressing the $k$-th pose of the assembly $A_i$ stably placed on the table. 
To determine the homogenous matrix $T_{ik}$, we need the information on 1) the grid point at which the assembly $A_i$ is placed, 2) the facet of the convex hull contacting the table, and 3) rotation of $A_i$ about the table normal. 
According to these information, we consider multiple candidates of the assembly $A_i$'s placing pose when planning the assembly motion of a robot. 

\subsection{Grasping Posture Set}


Let us consider a situation where the assembly $A_i$ is stably placed at one of the grid points. 
Let us also consider grasping the assembly $A_i$ by using the gripper $H_j$. 
If the assembly $A_i$ is composed of the parts $P_u, P_v, \cdots, P_w$, a set of grasping postures grasped by the gripper $H_j$ are composed of the elements of the database 
$G_{uj}, G_{vj}, \cdots, G_{wj}$. For each grasping posture, we consider solving the IK and checking the collision between the robot and the environment. 
We can obtain a set of IK solvable and collision free database of grasping postures 
$\hat{G}_{ijk}=(\hat{G}_{ijk1}, \cdots, \hat{G}_{ijkl})$ of the assembly $A_i$ grasped by the gripper $H_j$ where the assembly $A_i$ is stably placed at one of the grid points on the table.  


\subsection{Assembly Graph Search}

To plan the motion of a robot to assemble a product, we first search for the AND/OR graph. 
Then, by using the solution path of the AND/OR graph, we consider constructing the assembly graph where, by searching for the assembly graph, we can generate the motion of a robot to assemble a product. 
If we failed to find a path of the assembly graph, then we try to find another solution path of the AND/OR graph. 

\begin{figure}[t]
\centering
 	\includegraphics[width=85mm]{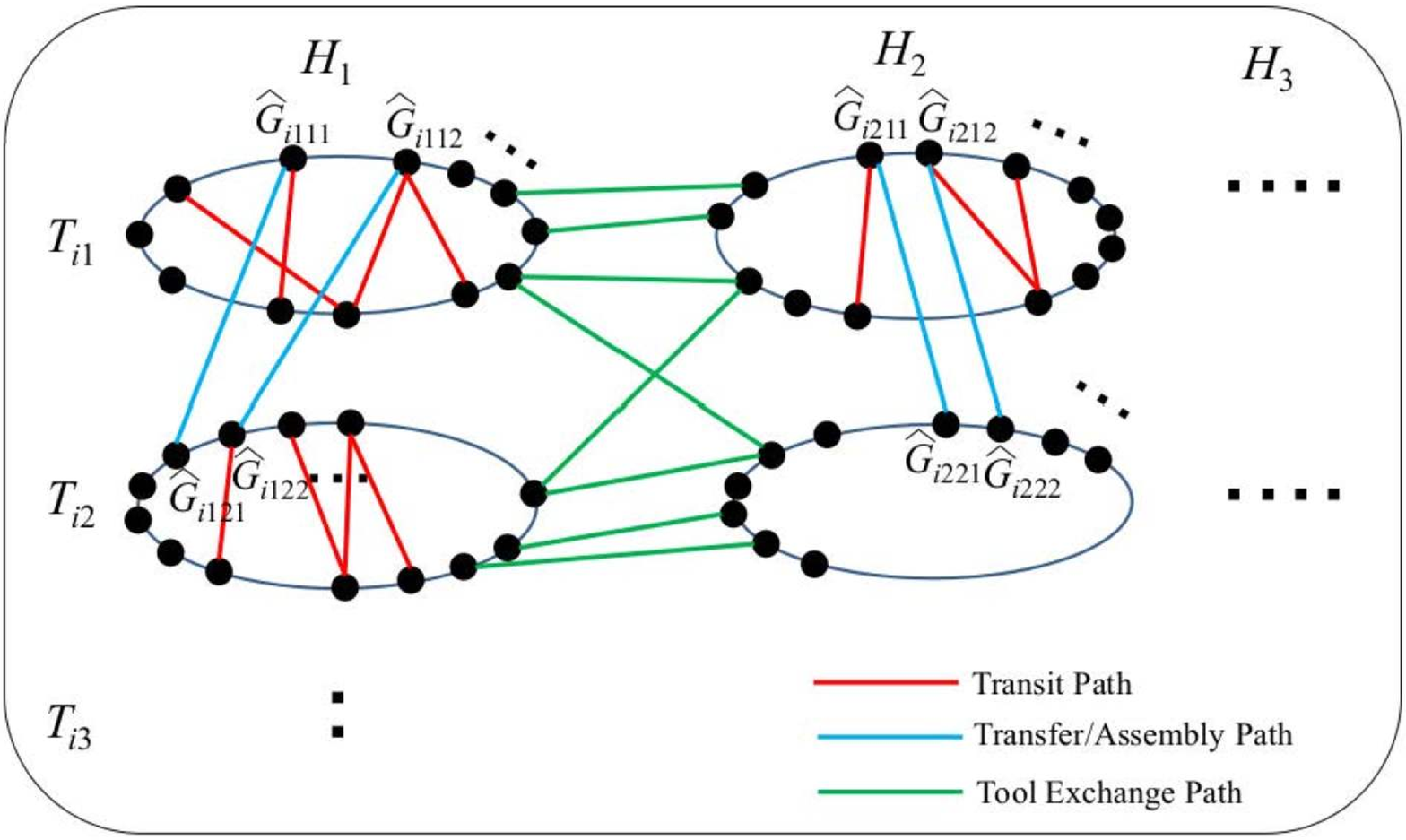}\\
 	\includegraphics[width=55mm]{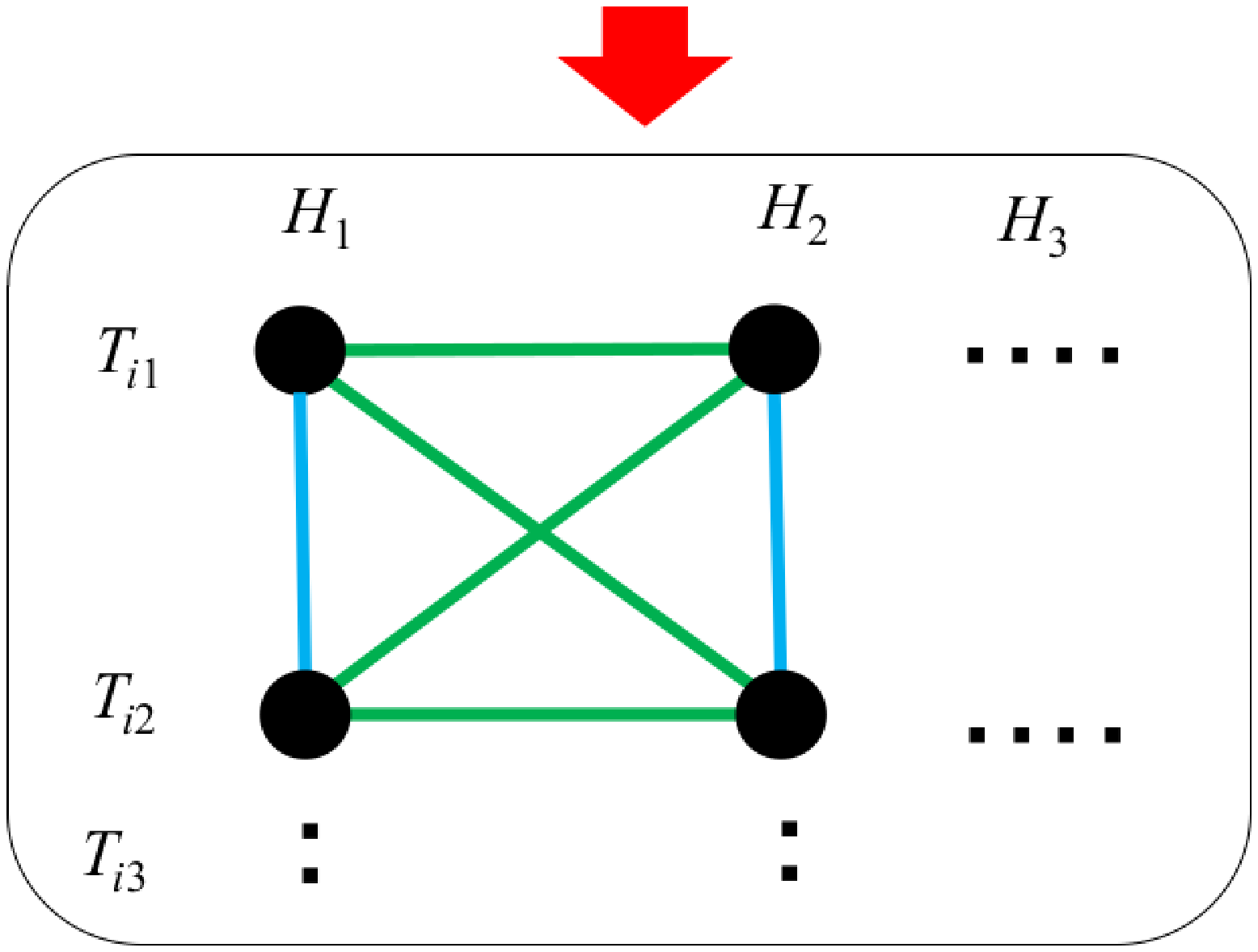}
	\caption{Path definitions of assembly graph and its simplified expression}
	\label{fig:graspPos-2}
\end{figure}

It would be easier for us to understand the structure of the assembly graph if we visualize it by drawing a circle for each $\hat{G}_{ijk}$ and plot the dots on the edge of the circle corresponding to 
$\hat{G}_{ijk1}, \cdots, \hat{G}_{ijkl}$ (Fig. \ref{fig:graspPos-2}). 

By extending the transit/transfer paths which have been introduced in manipulation planners such as\cite{Simeon}, we define the following three kinds of edges included in the assembly graph:

\noindent
\begin{description}
\item[{\it Transit Path}:] \mbox{}\\
Connect two nodes having the same object placing pose and having the same gripper but having different grasping pose. 
\item[{\it Transfer/assembly Path}:] \mbox{} \\
Connect two nodes having the same gripper and having the same grasping pose but having different object placing pose. 
\item[{\it Tool Exchange Path}:] \mbox{} \\
Connect two nodes having different gripper. 
\end{description}

\noindent
This visualization method is outlined in the upper side of Fig. \ref{fig:graspPos-2}. 
The transit paths can be expressed as edges connecting two nodes included in the same circle. 
The transfer/assembly path can be expressed as edges connecting two nodes included in the different circle but having the same gripper. 
The tool exchange path can be expressed by edges connecting two nodes included in the different circle, and having different grippers. 

Here, we consider the simplified assembly graph as shown in the lower side of Fig. \ref{fig:graspPos-2}. 
In this expression, each circle of the original manipulation graph is expressed as a single dot. 
Multiple transfer/assembly paths between two circles are merged into a single bold line. Multiple tool exchange paths between two circles are also merged into a single bold line. 
This simplified assembly graph does not explicitly include the transit paths. 

Next, we consider constructing the assembly graph. 
Since we imposed the assumptions 
{\bf A1}, {\bf A2} and {\bf A3}, we consider introducing the following four kinds of nodes included in the assembly graph:

\noindent
\begin{description}
\item[{\it Base Node}:] \mbox{} \\
An assembly of parts is placed at the assembly point. 
\item[{\it Assembly Node}:] \mbox{} \\
An assembly of parts is fit to another assembly of parts placed at the assembly point. 
\item[{\it Escape Node}:] \mbox{} \\
An assembly of parts is moved to one of the grid points included in the escape area. 
\item[{\it Initial Node}:] \mbox{} \\
A part is placed at the initial position. 
\end{description}

\noindent
Here, for an assembly of parts placed at one of the grid points, we can assume multiple nodes of the assembly graph depending on the rotation of the assembly about the table normal, multiple grasping configurations of the assembly, and multiple grippers grasping the assembly of parts. 

\begin{figure}[t]
\centering
 	\includegraphics[width=85mm]{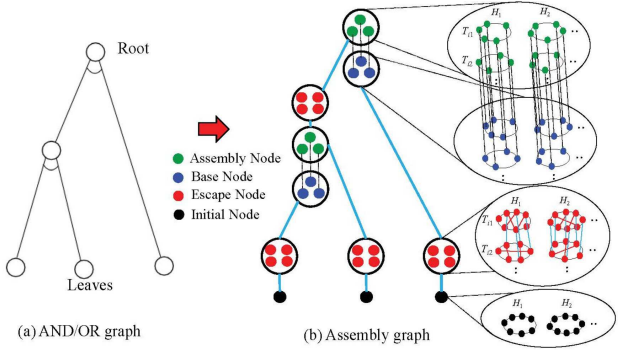}
	\caption{Transformation from AND/OR graph to assembly graph}
	\label{fig:agraph}
\end{figure}

\noindent
To construct the assembly graph from the AND/OR graph, the nodes of the AND/OR graph is replaced by a set of nodes of the assembly graph by the following rules: 
\begin{itemize}
\item The root node of the AND/OR graph is replaced by a set of base and assembly nodes (Fig. \ref{fig:agraph}) where one of the base nodes and one of the assembly nodes are connected by using the transfer/assembly path. 
\item The leaf nodes of the AND/OR graph are replaced by the initial and a set of the escape nodes where one of the initial nodes and one of the escape nodes are connected by using a transfer/assembly path and where the escape nodes are connected each other by using a transit and transfer paths. 
\item The nodes except for the root and the leaves are replaced by a set of base, assembly and escape nodes where one of the assembly nodes and one of the escape nodes are connected by using a transfer/assembly path, and where the escape nodes are connected each other by using a transit and transfer paths. 
\end{itemize}

\noindent
Here, the assembly nodes are automatically determined by the corresponding base nodes since assembly of parts defined in an assembly node includes the assembly of parts defined in the base node (shown in the dotted line in Fig. \ref{fig:agraph}). 

Then, we show a method for searching the assembly graph. 
In our method, we first search for a solution path of the simplified assembly graph. 
There are multiple root nodes included in the simplified assembly graph. 
From each root node, we consider searching for a solution path of the simplified assembly graph by using Dikstra method. 
Then, we consider selecting a root node where the path cost becomes minimum. 

For a solution path of the simplified assembly graph, we consider obtaining a sequence of assembly as will be explained in the next subsection. 

For a sequence of assembly tasks, we consider determining the grasping configuration. 
We first disregard the transit path and try to find grasping configurations with maximum grasping stability index \cite{icra13}. 
Our grasp stability index proposed in \cite{icra13} evaluates the contact area and can be applied for the soft-finger contact model. 
Thus, the gripper having fingers with large contact area tends to be selected. 

After obtaining a sequence of assembly, we check whether or not each edge included in the sequence is a feasible one by using the RRT (Rapidly-exploring Random Tree) algorithm. 
If RRT algorithm does not find a solution, we consider cutting the corresponding edge and try to find the grasping configuration again. 

\subsection{Assembly Sequence}

To perform an product assembly under the assumptions {\bf A1}, {\bf A2} and {\bf A3}, an assembly of parts has to be first placed at the assembly point before it is fit to another assembly of parts. 
Let us consider the case where a robot assembles the assembly $A_u$ to the assembly $A_v$. 
After a robot places the assembly $A_u$ to the assembly point, a robot may first exchange the gripper, then picks the assembly $A_v$, and finally assembles it to the assembly $A_u$. 
However, the solution of the assembly graph obtained in the previous subsection does not include such information. 
In this subsection, we consider generating an assembly sequence taking the exchange of grippers into consideration. 
From the assembly graph constructed in the previous subsection, a sequence of assembly is generated by using the following method:

\begin{enumerate}
\item Push the root node of the assembly graph to the stack. 
\item Iterate the following steps until the stack becomes empty
 \begin{enumerate}
 \item Pop the stacked nodes.  Connect the path including an assembly node between the stacked node and either an initial or an escape node to the solution path by using the tool exchange path. 
 \item If the last node of the solution path is an escape node, push the escape node to the stack. 
 \item Connect the path between the corresponding base node and either an initial or an escape node to the solution path by using the tool exchange path. 
 \item If the last node of the solution path is an escape node, push the escape node to the stack. 
 \end{enumerate}
\end{enumerate}

\begin{figure}[t]
\centering
 	\includegraphics[width=85mm]{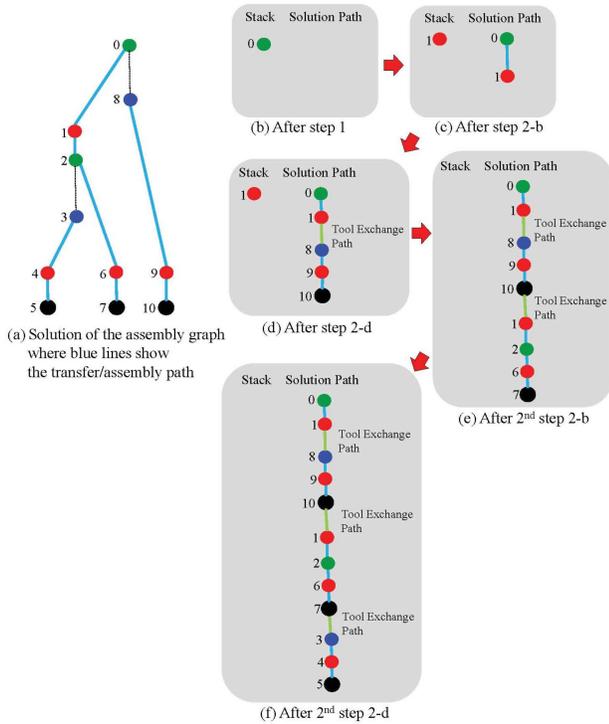}
	\caption{Searcning algorithm of assembly sequence}
	\label{fig:search}
\end{figure}

\noindent
Fig. \ref{fig:search} shows how the algorithm shown in the example of Fig. \ref{fig:agraph} works. 
As shown in this figure, the base node is scheduled before the assembly node and whole assembly sequence can be performed where an adequate gripper is selected according to the assembly of parts. 

\section{Results}

In this section, we show some numerical examples to show the effectiveness of our proposed method. 
We prepared two two-fingered parallel grippers used to assemble a product. 
One of the grippers has relatively small contact area where it would be suitable 
for assembling a small object. 
On the other hand, the other gripper has relatively large contact area where it would be 
suitable for assembling a large object. 

\begin{figure}[t]
\centering
 	\includegraphics[width=50mm]{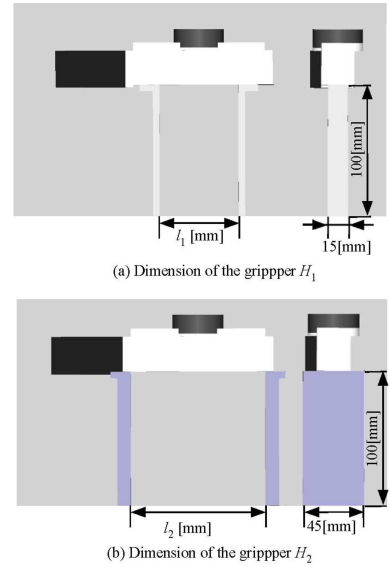}
	\caption{Two fingered parallel grippers used in numerical examples}
	\label{fig:gripper}
\end{figure}

\begin{figure}[t]
\centering
 	\includegraphics[width=50mm]{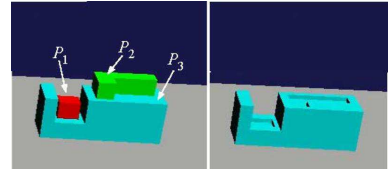}
	\caption{Assembled product used in Example 1}
	\label{fig:example1}
\end{figure}

\begin{figure}[t]
\centering
 	\includegraphics[width=83mm]{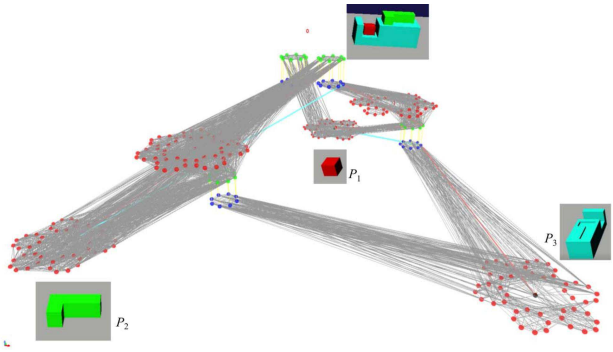}
	\caption{Assembly graph generated in Example 1}
	\label{fig:graph1}
\end{figure}

\begin{figure}[t]
\centering
 	\includegraphics[width=83mm]{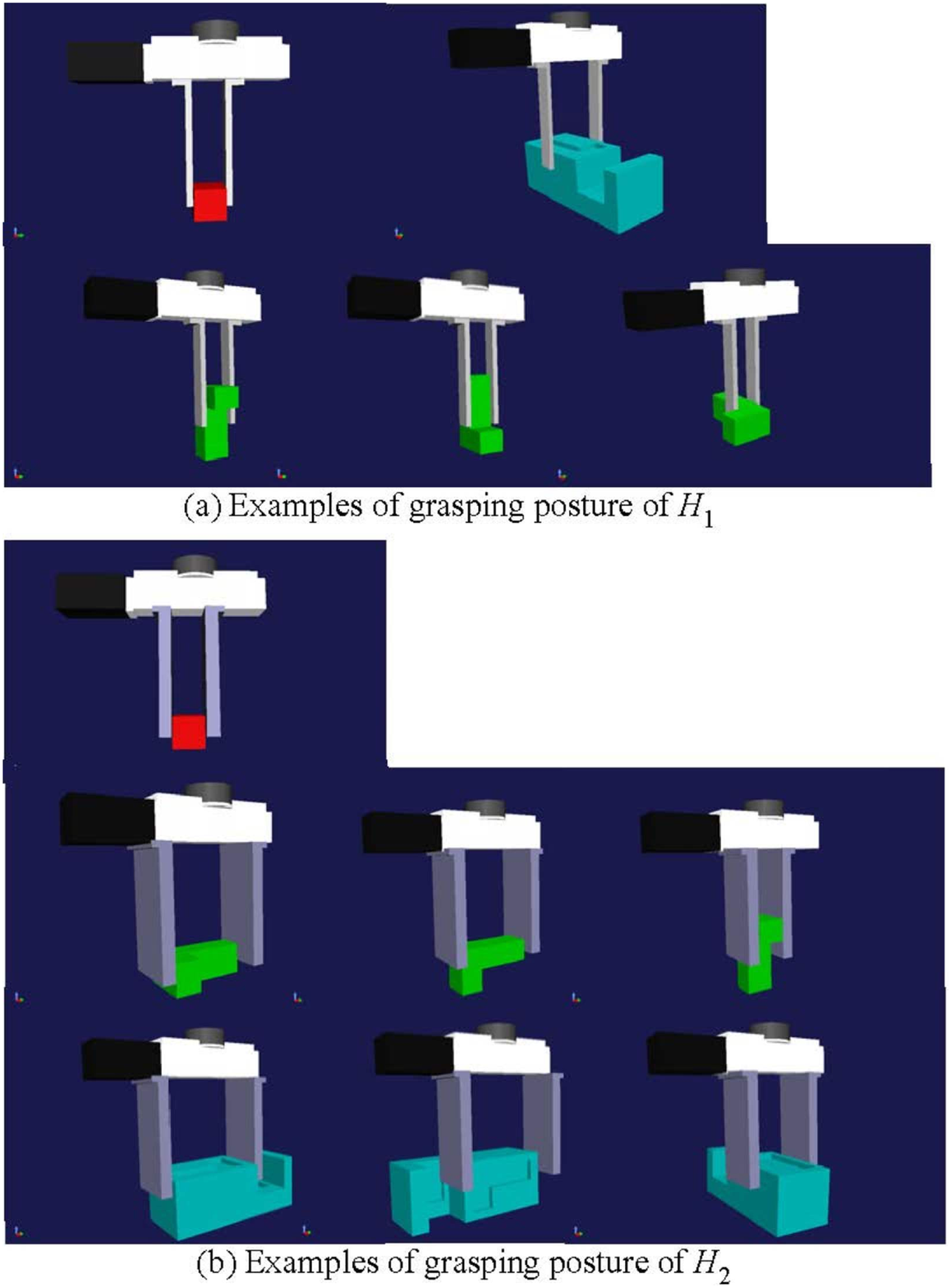}
	\caption{Grasping posture used in Example 1}
	\label{fig:grasp1}
\end{figure}

\begin{figure}[t]
\centering
 	\includegraphics[width=83mm]{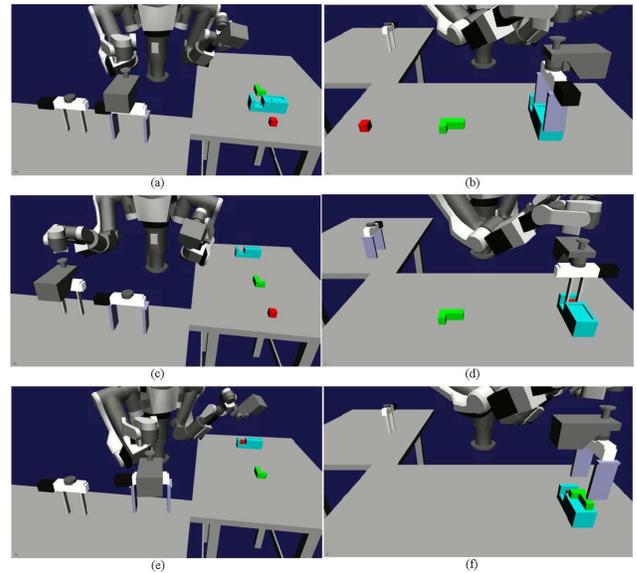}
	\caption{Snapshot of assembly motion of Example 1}
	\label{fig:motion1}
\end{figure}

In the first example, the robot tries to assemble a product made of three blocks as shown in 
Fig. \ref{fig:example1}. In this example, the width $l_1$ and $l_2$ of the gripper shown in Fig. \ref{fig:gripper} 
is set as $0 \le l_1 \le 0.06${[m]} and $0 \le l_2 \le 0.1${[m]}, 
respectively. 
The simplified assembly graph is shown in Fig.\ref{fig:graph1} where the solution path is expressed by the red transfer paths 
and blue tool-exchange paths. We used a grasp planner proposed in \cite{Harada1} to calculate the grasping posture. The number of grasping posture included in the database is 
${\rm dim}(G_{11})=347$, ${\rm dim}(G_{12})=443$, 
${\rm dim}(G_{21})=1221$, ${\rm dim}(G_{22})=2162$, 
${\rm dim}(G_{31})=12$, and ${\rm dim}(G_{32})=18$. 
Some examples of grasping postures are shown in Fig. \ref{fig:grasp1}. 
It took about 2 {[min]} to calculate the solution path by using 
the 3.4{[GHz]} Quad-core PC. Some examples of grasping posture are shown in Fig. \ref{fig:grasp1}. 
The motion of the robot is shown in Fig. \ref{fig:motion1} where the robot first used the hand $H_2$ to stably pick the part $P_3$. 
Then, the robot used the hand $H_1$ to pick the part 
$P_1$ and assembled it to the part $P_3$. Here, it is impossible to 
use the large hand $H_2$ to assemble $P_1$ 
to the concaved part of $P_3$. Finally, the robot used again the hand 
$H_2$ to assemble the part $P_2$. 

\begin{figure}[t]
\centering
 	\includegraphics[width=83mm]{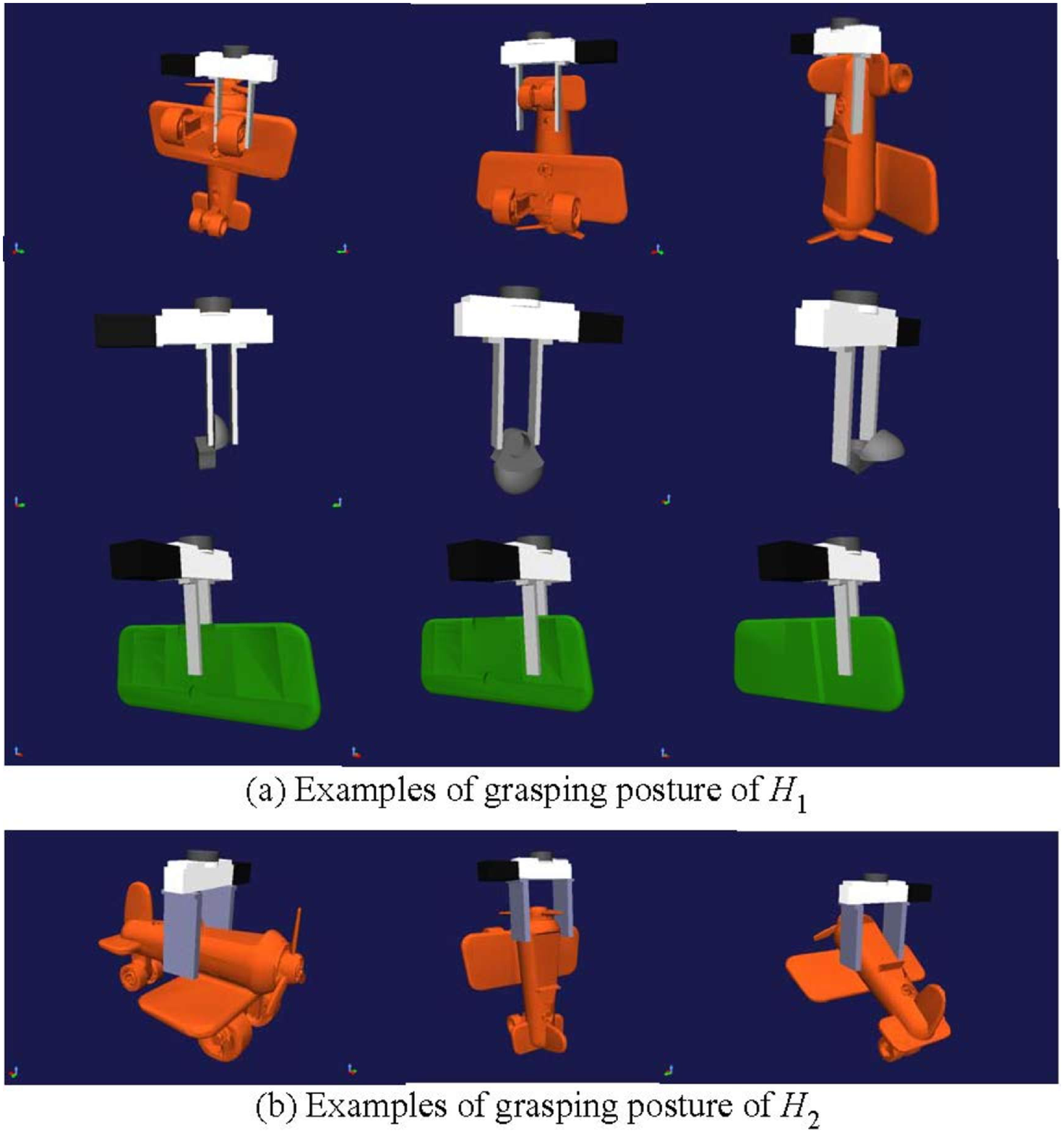}
	\caption{Grasping posture used in Example 2}
	\label{fig:grasp2}
\end{figure}

\begin{figure}[t]
\centering
 	\includegraphics[width=83mm]{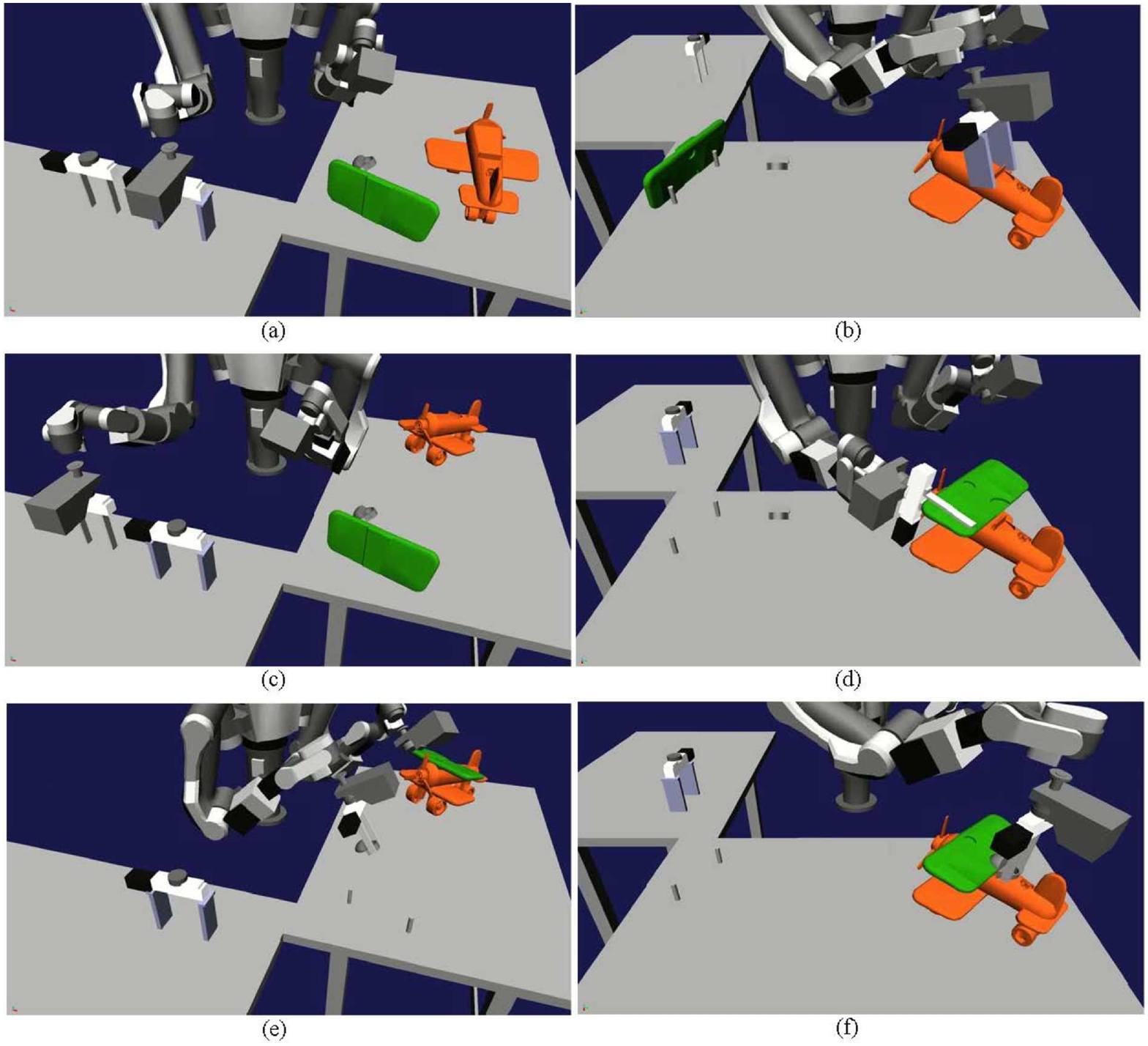}
	\caption{Snapshot of assembly motion of Example 2}
	\label{fig:motion2}
\end{figure}

In the second example, we consider the assembly of a toy airplane where 
its AND/OR graph is shown in Fig. \ref{fig:andor}. In this example, 
we consider the assembly problem of three parts: 
$A_2$, $A_5$ and $A_{11}$. 
The width $l_1$ and $l_2$ of the gripper shown in Fig. \ref{fig:gripper} 
is set as $0 \le l_1 \le 0.06${[m]} and $0.4 \le l_2 \le 0.1${[m]}, 
respectively. In this example, the gripper $H_1$ is suitable for grasping a thin object while 
the gripper $H_2$ is suitable for grasping a thick one. 
The number of grasping posture included in the database is 
${\rm dim}(G_{11})=49$, ${\rm dim}(G_{12})=58$, 
${\rm dim}(G_{21})=4$, ${\rm dim}(G_{22})=24$, 
${\rm dim}(G_{31})=0$, and ${\rm dim}(G_{32})=263$. 
Some examples of grasping postures are shown in Fig. \ref{fig:grasp2}. 
It took 28{[sec]} to plan the robot motion. 
The motion of the robot is shown in Fig. \ref{fig:motion2} where the robot first used the 
hand $H_2$ to stably pick the thick $A_{11}$. 
Then, the robot used the hand $H_1$ to pick the part 
$A_2$ and $A_5$. Since the same hand is used two individual assembly tasks sequentially connected, 
the robot does not exchange the gripper. 

\begin{figure}[t]
\centering
 	\includegraphics[width=50mm]{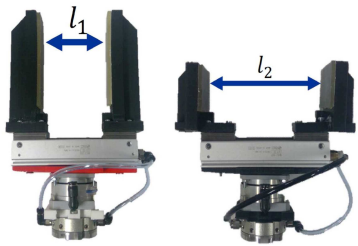}
	\caption{Grippers used for experiment}
	\label{fig:grippers}
\end{figure}

\begin{figure}[t]
\centering
 	\includegraphics[width=83mm]{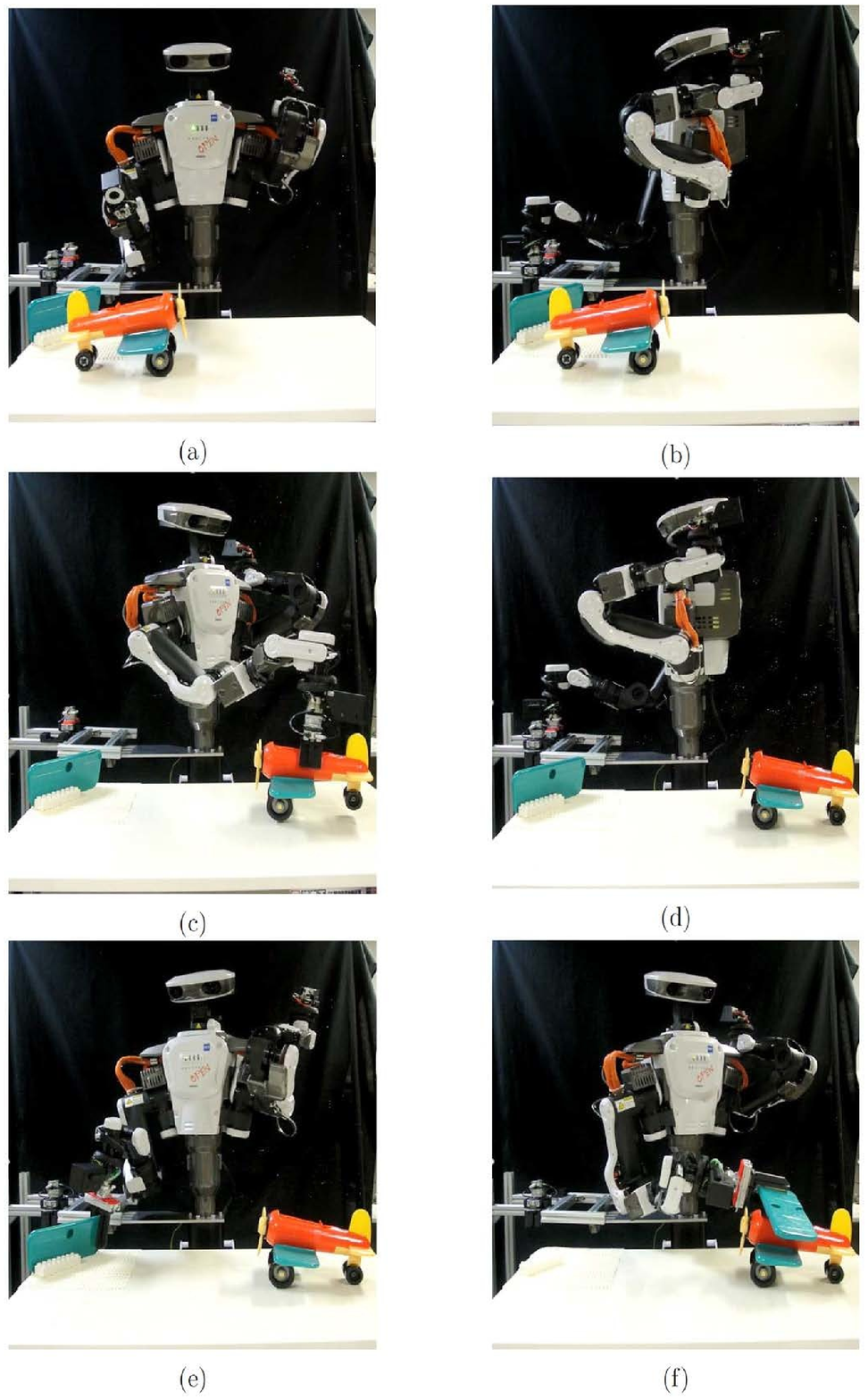}
	\caption{Snapshot of experiment}
	\label{fig:experiment}
\end{figure}

Finally, we performed experiment on the toy airplane assembly. 
In this experiment, we used two kinds of parallel jaw gripper as shown in Fig. \ref{fig:grippers} corresponding to the simulation result 
of toy airplane assembly. 
Fig. \ref{fig:experiment} shows experimental result where a robot stably grasps each parts and successfully conducted the assembly task. 

\section{Conclusion}

In this paper, we proposed a grasp/assembly planner for a manipulator 
which can simultaneously plan assembly sequence, robot motion, grasping configuration, and 
exchange of grippers. 
For a given AND/OR graph of an assembly task, we generated the assembly graph where 
its edges are composed of three kinds of paths, i.e., the transfer/assembly path, the transit path 
and the tool exchange path. 
We showed numerical examples assuming two kinds of two-fingered parallel grippers 
where one of the grippers is suitable for grasping a small part and the other is suitable for grasping 
a large part. 

For a future research, we consider conducting a real world experiment. Motion optimization is also considered 
to be our future research topic. 





\end{document}